\documentclass{article}
\usepackage[utf8]{inputenc}

\title{Efficient Action Poisoning Attacks on Linear Contextual Bandits}
\author{Guanlin Liu, Lifeng Lai\\
Department of ECE, University of California, Davis\\
	Email:\{glnliu,lflai\}@ucdavis.edu}

\usepackage{PRIMEarxiv}
\usepackage{amsthm}
\newtheorem{assump}{Assumption}
\newtheorem{proposition}{Proposition}

\newtheorem{theorem}{Theorem}

\newtheorem{lemma}{Lemma}

\usepackage{bbm}        
\usepackage{amsmath}
\usepackage{comment}
\usepackage{amssymb}
\usepackage{subfigure}
\usepackage{graphicx} 
\usepackage{algorithm}
\usepackage{algorithmic}

\begin{document}

\maketitle

\begin{abstract}
Contextual bandit algorithms have many applicants in a variety of scenarios. %such as displaying advertisements,  recommender systems, virtual reality, etc. 
In order to develop trustworthy contextual bandit systems, understanding the impacts of various adversarial attacks on contextual bandit algorithms is essential. %for the safe applications of these models. Recent works show that some adversary can force the no-regret contextual bandit algorithms into a desired behavior and expose a significant security threat to contextual bandit algorithms. However, prior works on adversarial attacks against contextual bandit algorithms mainly focus on the reward poisoning attacks or the context poisoning attacks. 
In this paper, we propose a new class of attacks: action poisoning attacks, where an adversary can change the action signal selected by the agent. We design action poisoning attack schemes against linear contextual bandit algorithms in both white-box and black-box settings. We further analyze the cost of the  proposed attack strategies for a very popular and widely used bandit algorithm: LinUCB. We show that, in both white-box and black-box settings, the proposed attack schemes can force the LinUCB agent to pull a target arm very frequently by spending only logarithm cost.
\end{abstract}

\section{Introduction}
Multiple armed bandits (MABs), a popular framework of sequential decision making model, has been widely investigated %in machine learning and artificial intelligence
and has many applicants in a variety of scenarios \cite{Chapelle:2014:SSR:2699158.2532128,Lai:TMC:11,kveton2015cascading}. The contextual bandits model is an extension of the multi-armed bandits model with contextual information. At each round, the reward is associated with both the arm (a.k.a, action) and the context, while the reward of stochastic MABs is only associated with the arm. Contextual bandits algorithms have a broad range of applications, such as recommender systems~\cite{li2010contextual}, wireless networks~\cite{saxena2019contextual}, etc.

In the modern industry-scale applications of bandit algorithms, action decisions, reward signal collection, and policy iterations are normally implemented in a distributed network. Action decisions and reward signals may need to be transmitted over communication links. When data packets containing the reward signals and action decisions etc are transmitted through the network, the adversary can implement adversarial attacks by intercepting and modifying these data packets. As the result, poisoning attacks on contextual bandits could possibly happen. In many applications of contextual bandits, an adversary may have an incentive to force the contextual bandits system to learn a specific policy. For example, a restaurant may attack the bandit systems to force the systems into increasing the restaurant's exposure. Thus, understanding the risks of different kinds of adversarial attacks on contextual bandits is essential for the safe applications of the contextual bandit model and designing robust contextual bandit systems. 

While there are many existing works addressing adversarial attacks on supervised learning models \cite{42503, moosavi2017universal, cohen2019certified, dohmatob19nofreelunch, wang2019convergence, dasgupta19Teaching, cicalese2020teaching}, the understanding of adversarial attacks on contextual bandit models is less complete. Of particular relevance to our work is a line of interesting recent work on adversarial attacks on MABs~\cite{jun2018adversarial,liu2019data, AAonB} and on linear contextual bandits~\cite{ma2018data, AttacksLinearBandits}. 
In recent works in MABs setting, the types of attacks include both reward poisoning attacks and action poisoning attacks. In the reward poisoning attacks, there is an adversary who can manipulate the reward signal received by the agent~\cite{jun2018adversarial, liu2019data}. In the action poisoning attacks, the adversary can manipulate the action signal chosen by the agent before the environment receives it~\cite{AAonB}. Among existing works on adversarial attacks against linear contextual bandits, they focus on the reward poisoning~\cite{ma2018data, AttacksLinearBandits} or context poisoning attacks~\cite{AttacksLinearBandits}. In the context poisoning attacks, the adversary can modify the context observed by the agent without changing the reward associated with the context. There are also some recent interesting work on adversarial attacks against reinforcement learning algorithms under various setting~\cite{2017vulnerability, 2019deceptive, PolicyPoisoning, Adaptive-Reward-Poisoning, ICLR2021, policyteaching, arxiv-2102-08492, liu2021provably}.

In this paper, we aim to investigate the impact of action poisoning attacks on contextual bandit models. To our best knowledge, this paper is the first work to analyze the impact of action poisoning attacks on contextual bandit models. More detailed comparisons of various types of attacks against contextual bandits will be provided in Section~\ref{sec:model}. We note that the goal of this paper is not to promote any particular type of poisoning attack. Rather, our goal is to understand the potential risks of action poisoning attacks. 
We note that for the safe applications and design of robust contextual bandit algorithms, it is essential to address all possible weaknesses of the models and understanding the risks of different kinds of adversarial attacks. Since the action poisoning attack is an important aspect of poisoning attacks and may threaten the bandit systems, it is important to understand the potential risks of action poisoning attacks. %is important for the safe applications of contextual bandit models.

In this paper, we study the action poisoning attack against linear contextual bandit in both white-box and black-box settings. In the white-box setting, we assume that the attacker knows the coefficient vectors associated with arms. Thus, at each round, the attacker knows the mean rewards of all arms. While it is often unrealistic to exactly know the coefficient vectors, the understanding of the white-box attacks could provide valuable insights on how to design the more practical black-box attacks. In the black-box setting, we assume that the attacker has no prior information about the arms and does not know the agent's algorithm. The limited information that the attacker has are the context information, the action signal chosen by the agent, and the reward signal generated from the environment. In both white-box and black-box settings, the attacker aims to manipulate the agent into frequently pulling a target arm chosen by the attacker with a minimum cost. The cost is measured by the number of rounds that the attacker changes the actions selected by the agent. The contributions of this paper are:

\begin{itemize}
    \item We propose a new online action poisoning attack against contextual bandit in which the attacker aims to force the agent to frequently pull a target arm chosen by the attacker via strategically changing the agent's actions. %We compared the action poisoning attacks with the reward action attacks and the context action attacks.
    \item We introduce a white-box attack strategy that can manipulate any sublinear-regret linear contextual bandit agent into pulling a target arm $T-o(T)$ rounds over a horizon of $T$ rounds, while incurring a cost that is sublinear dependent on $T$.
    \item We design a black-box attack strategy whose performance nearly matches that of the white-box attack strategy. We apply the black-box attack strategy against a very popular and widely used bandit algorithm: LinUCB. We show that our proposed attack scheme can force the LinUCB agent into pulling a target arm $T-O(\log^3 T)$ times with attack cost scaling as $O(\log^3 T)$.
    \item We evaluate our attack strategies using both synthetic and real-world datasets. We observe empirically that the total cost of our black-box attack is sublinear for a variety of contextual bandit algorithms. %Our work exposes a significant security threat of the action poisoning attacks on linear contextual bandits.
\end{itemize}

\section{Related Work}

In this section, we discuss related works on two parts: adversarial attacks that cause standard bandit algorithms to
fail and robust bandit algorithms that can defend against such attacks.

\noindent\textbf{Attacks Models.} In MABs setting,~\cite{jun2018adversarial} proposes an interesting reward poisoning attack strategy that can force $\epsilon$-Greedy or upper confidence bound (UCB) agent to select a target arm while only spending logarithmic effort. The main idea of the attack scheme in~\cite{jun2018adversarial} is to modify the reward signals associated with non-target arms to smaller values. As the agent only observes the modified reward signals, the target arm appears to the optimal arm for the agent.
\cite{liu2019data} proposes an optimization based framework for offline reward poisoning attacks on MABs. Furthermore, it studies the online attacks on MABs, and proposes an adaptive attack strategy that is effective in attacking any bandit algorithm without knowing what particular algorithm the agent is using. \cite{AAonB} proposes an adaptive action poisoning attack strategy that can force the UCB agent to pull a target arm $T-O(\log T)$ times over $T$ rounds while the total attack cost being only $O(\log T)$. 

In linear contextual bandit setting, \cite{ma2018data} studies offline reward poisoning attacks and investigates the feasibility and the impacts of such attacks. The attacker in~\cite{ma2018data} aims to force the agent to pull a target arm on a particular context. \cite{AttacksLinearBandits} extends the attack idea of~\cite{jun2018adversarial, liu2019data} to linear contextual bandits. It proves that the proposed reward poisoning attack strategy can force any bandit algorithms to pull a specific set of arms when the rewards are bounded. It introduces an adaptive reward poisoning attack strategy and observes empirically that the total cost of the adaptive attack is sublinear. In addition, \cite{AttacksLinearBandits} analyzes the context poisoning attacks in white-box setting and shows that LinUCB is vulnerable to such attack. In the filed of adversarial attacks on RL, \cite{liu2021provably} studies black-box action poisoning attacks against RL.

\noindent\textbf{Robust algorithms.} Lots of efforts have been made to design robust bandit algorithms to defend adversarial attacks. 
In the MABs setting, \cite{lykouris2018stochastic} introduces a bandit algorithm, called Multilayer Active Arm Elimination Race algorithm, that is robust to reward poisoning attacks by using a multi-layer approach. 
\cite{gupta2019better} presents an algorithm named BARBAR that is robust to reward poisoning attacks and the regret of the proposed algorithm is nearly optimal. %BARBAR is similar in spirit to the Active Arm Elimination algorithm. However, BARBAR never eliminates an arm permanently. 
\cite{Guan:AAAI} considers a reward poisoning attack model where an adversary attacks with a certain probability at each round. As its attack value at each round can be arbitrary and unbounded, the attack model could be powerful. The paper proposes algorithms that are robust to these types of attacks.
\cite{feng2020intrinsic} introduces a reward poisoning attack setting where each arm can only manipulate its own reward. Every arm can be considered as an adversary and each arm seeks to maximize its own expected number of pull count. Under this setting, \cite{feng2020intrinsic} proves that Thompson Sampling, UCB, and $\epsilon$-greedy can be modified to be robust to such attacks.% and achieve a regret upper bound that increases over rounds in a logarithmic order or increases with attack cost in a linear order. 
\cite{AA&DonB} introduce a bandit algorithm, called MOUCB, that is robust to action poisoning attacks and achieves a regret upper bound that increases over rounds in a logarithmic order or increases with attack cost in a linear order. 

In the linear contextual bandit setting, \cite{Bogunovic2021robust} proposes a stochastic linear bandit algorithm, called Robust Phased Elimination (RPE), that is robust to reward poisoning attacks. It provides two variants of RPE algorithm which separately work on known attack budget case and agnostic attack budget case.
\cite{ding2021robust} provides a robust linear contextual bandit algorithm, called RobustBandit, that works under both the reward poisoning attacks and context poisoning attacks.

\section{ Problem Setup} \label{sec:model} 
%In this paper, we consider the stochastic linear contextual bandit with disjoint linear models where different arms are associated with different coefficient vectors. We introduce the action poisoning attacks against the linear contextual bandit algorithms.

%\subsection{Review of linear contextual bandit}
Consider the standard contextual linear bandit model in which the environment consists of $K$ arms. In each round $t = 1, 2, 3, \dots, T$, the agent observes a context $x_t \in \mathbb{R}^d$, pulls an arm $I_t$ and receives a reward $r_{t,I_t}$. Each arm $i$ is associated with an unknown but fixed coefficient vector $\theta_i \in \mathbb{R}^d$.  In each round $t$, the reward is $$r_{t,I_t} = \langle x_t,\theta_{I_t} \rangle + \eta_t, $$ where $\eta_t$ is a conditionally independent zero-mean $R$-subgaussian noise and $\langle \cdot,\cdot \rangle$ denotes the inner product.
Hence, the expected reward of arm $i$ under context $x_t$ follows the linear setting:
\begin{equation}
\mathbb{E}[r_{t,i}]=\langle x_t,\theta_{i} \rangle
\end{equation}
for all $t$ and all arm $i$. If we consider the $\sigma$-algebra $F_t= \sigma(x_1,\dots,x_{t+1},\eta_1,\dots,\eta_t)$, $x_t$ becomes $F_{t-1}$ measurable and $\eta_t$ becomes $F_{t}$ measurable.

In this paper, we assume that there exist $L > 0$ and $S > 0$, such that for all round $t$ and arm $i$, $\Vert x_t \Vert_2 \le L$ and $\Vert \theta_i \Vert_2 \le S$, where $\Vert\cdot\Vert_2$ denotes the $\ell_2$-norm. 
We assume that there exist $\mathcal{D} \subset \mathbb{R}^d$ such that for all $t$, $x_t \in \mathcal{D}$ and, for all $x \in \mathcal{D}$ and all arm $i$, $\langle x,\theta_i \rangle > 0$.

The agent is interested in minimizing the cumulative pseudo-regret%, and the cumulative pseudo-regret can be formally written as 
\begin{equation}
    \bar{R}_T=\sum_{t=1}^T \left(\langle x_t,\theta_{I_t^*} \rangle - \langle x_t,\theta_{I_t} \rangle \right),
\end{equation}
where $I_t^* = \arg \max_i \langle x_t,\theta_i \rangle$.

%\subsection{Action poisoning attack} \label{sec:apaintro}
In this paper, we introduce a novel adversary setting, in which the attacker can manipulate the action chosen by the agent. In particular, at each round $t$, after the agent chooses an arm $I_t$, the attacker can manipulate the agent's action by changing $I_t$ to another $I_t^0 \in \{1, \dots, K\}$. If the attacker decides not to attack, $I_t^0=I_t$. The environment generates a random reward $r_{t,I_t^0}$ based on the post-attack arm $I_t^0$ and the context $x_t$. Then the agent and the attacker receive reward $r_{t,I_t^0}$ from the environment. Since the agent does not know the attacker's manipulations and the presence of the attacker, the agent will still view  $r_{t,I_t^0}$ as the reward corresponding to the arm $I_t$. %The action poisoning attack model is summarized in Algorithm~\ref{alg:attackmodel} .

%\begin{algorithm}[htb] 
%\caption{Action poisoning attacks on contextual linear bandit agent} 
%\label{alg:attackmodel} 
%\begin{algorithmic}[1] 
%\FOR{$t = 1, 2, \dots, T$}
%\STATE Agent chooses arm $I_t$ after observing the context $x_t$.\
%\STATE Attacker observes the agent's action $I_t$. If the attacker decides to attack, it manipulates the action to $I_t^0$. If the attacker does not attack, $I_t^0 = I_t$.
%\STATE The environment generates reward $r_{t, I_t^0}$ according to arm $I_t^0$ and context $x_t$.
%\STATE The agent and attacker receive reward $r_{t, I_t^0}$.
%\ENDFOR
%\end{algorithmic}
%\end{algorithm}

The goal of the attacker is to design an attack strategy so as to manipulate the agent into pulling a target arm very frequently but by making attacks as rarely as possible. Without loss of generality and for notation convenience, we assume arm $K$ is the ``attack target" arm or target arm. Define the set of rounds when the attacker decides to attack as $\mathcal{C}:=\{ t: t\le T,I_t^0\not= I_t \}$. The cumulative attack cost is the total number of rounds where the attacker decides to attack, i.e., $|\mathcal{C}|$. The attacker can monitor the contexts, the actions of the agent and the reward signals from the environment. %Furthermore, the attacker can introduce action poisoning attacks.

We now compare the three types of poisoning attacks against contextual linear bandit: reward poisoning attack, action poisoning attack and context poisoning attack. In the reward poisoning attack~\cite{ma2018data, AttacksLinearBandits}, after the agent observes context $x_t$ and chooses arm $I_t$, the environment will generate reward $r_{t,I_t}$ based on context $x_t$ and arm $I_t$. Then, the attacker can change the reward $r_{t,I_t}$ to $\widetilde{r}_t$ and feed $\widetilde{r}_t$ to the agent.
Compared with the reward poisoning attacks, the action poisoning attack considered in this paper is more difficult to carry out. In particular, as the action poisoning attack only changes the action, it can impact but does not have direct control of the reward signal. By changing the action $I_t$ to $I_t^0$, the reward received by the agent is changed from $r_{t,I_t}$ to $r_{t,I_t^0}$ which is a random variable drawn from a distribution based on the action $I_t^0$ and context $x_t$. This is in contrast to reward poisoning attacks where an attacker has direct control and can change the reward signal to any value $\widetilde{r}_t$ of his choice. 
 In the context poisoning attack~\cite{AttacksLinearBandits}, the attacker only changes the context shown to the agent. The reward is also generated based on the true context $x_t$ and the agent's action $I_t$. Nevertheless, the agent's action may be indirectly impacted by the manipulation of the context, and so as the reward. Since the attacker attacks before the agent pulls an arm, the context poisoning attack is the most difficult to carry out. As mentioned in the introduction, the goal of this paper is not to promote any particular types of poisoning attacks. Instead, our goal is to understand the potential risks of action poisoning attacks, as the safe applications and design of robust contextual bandit algorithm relies on the addressing all possible weakness of the models.% and understanding the risks of different kinds of adversarial attacks.

%\begin{figure}
%\centering
%\subfigure[Contextual bandit]{\includegraphics[width=3.8 cm]{withoutattack.png}} \hspace{6mm}
%\subfigure[Poison action]{\includegraphics[width=3.8 cm]{actionattack.png}}
%\\
%\centering
%\subfigure[Poison context]{\includegraphics[width=3.8 cm]{contextattack.png}} \hspace{6mm}
%\subfigure[Poison reward]{\includegraphics[width=3.8 cm]{rewardattack.png}}
%\caption{Poisoning attacks on contextual bandits}
%\end{figure}

As the action poisoning attack only changes the actions, it can impact but does not have direct control of the agent's observations. Furthermore, when the action space is discrete and finite, the ability of the action poisoning attacker is severely limited. It is reasonable to limit the choice of the target policy. Here we introduce an important assumption that the target arm is not the worst arm:

\begin{assump} \label{asp:nonworst}
 For all $x \in \mathcal{D}$, $\langle x,\theta_{K} \rangle > \min_{i \in [K]} \langle x,\theta_{i} \rangle$.
\end{assump}

\begin{figure} [t]
	\centering
	\includegraphics[width=0.4\textwidth]{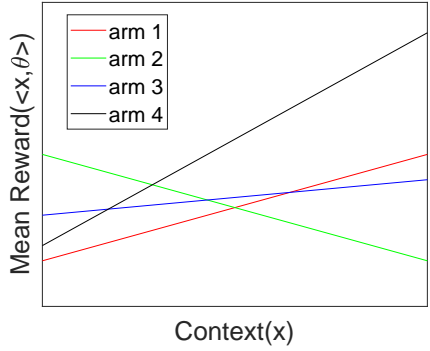}  %Reduce the figure size so that it is slightly narrower than the column.
	\caption{An example of one dimension linear contextual bandit model.}
	\label{fig:asp1}
\end{figure}

If the target arm is the worst arm in most contexts, the attacker should change the target arm to a better arm or the optimal arm so that the agent learns that the target set is optimal for almost every context. In this case, the cost of attack may be up to $O(T)$. Assumption~\ref{asp:nonworst} does not imply that the target arm is optimal at some contexts. The target arm could be sub-optimal for all contexts. Fig.~\ref{fig:asp1} shows an example of one dimension linear contextual bandit model, where the $x$-axis represents the contexts and the $y$-axis represents the mean rewards of arms under different contexts. As shown in Fig.~\ref{fig:asp1}, the arm 3 and arm 4 satisfy the assumption~\ref{asp:nonworst}. In addition, the arm 3 is not optimal at any context.

Under assumption~\ref{asp:nonworst}, there exists $\alpha \in ( 0, \frac{1}{2} )$ such that  $ \max_{x \in \mathcal{D}} \frac{\min_i \langle x,\theta_i \rangle}{\langle x,\theta_K \rangle} \le (1 - 2\alpha)$. Equivalently, Assumption~\ref{asp:nonworst} implies that there exists $ \alpha \in ( 0, \frac{1}{2} ) $, such that for all $t$, we have 
\begin{equation}
    (1 - 2 \alpha) \langle x_t,\theta_{K} \rangle \ge \min_{i \in [K]}\langle x_t,\theta_{i} \rangle.\label{eq:alpha}
\end{equation}

\section{Attack Schemes and Cost Analysis}
In this section, we introduce action poisoning attack schemes in the white-box setting and black-box setting respectively. In order to demonstrate the significant security threat of action poisoning attacks to linear contextual bandits, we investigate our action poisoning attack strategy against a widely used algorithm: LinUCB algorithm. Furthermore, we analyze the attack cost of our action poisoning attack schemes.

\subsection{Overview of LinUCB}
For reader's convenience, we first provide a brief overview of the LinUCB algorithm~\cite{li2010contextual}. 
%Several efficient algorithms have been proposed in the literature for the linear contextual bandit problems. Some contextual bandit algorithms, such as LinUCB, LinTS and OFUL
The LinUCB algorithm is summarized in Algorithm~\ref{alg:Framwork}.
The main steps of LinUCB are to obtain estimates of the unknown parameters $\theta_i$ using past observations and then make decisions based on these estimates. Define $\tau_i(t):=\{ s:s \le t, I_s=i \}$ as the set of rounds up to $t$ where the agent pulls arm $i$. Let $N_i(t)=|\tau_i(t)|$. Then, at round $t$, the $\ell_2$-regularized least-squares estimate of $\theta_i$ with regularization parameter $\lambda > 0$ is obtained by~\cite{li2010contextual}
\begin{equation} \label{ridge}
\hat{\theta}_{t,i}=V_{t,i}^{-1} \left(\sum_{k \in \tau_i(t-1)} r_{t,i} x_{k}\right),
\end{equation}
where $V_{t,i} = \sum_{k \in \tau_i(t-1)} x_{k} x_{k}^T + \lambda \mathbf{I}$ with $\mathbf{I}$ being identity matrix.

%In this paper, we considered the standard LinUCB with disjoint linear models. 

After $\hat{\theta}_i$'s are obtained, at each round, an upper confidence bound of the mean reward has to be calculated for each arm (step 5 of Algorithm~\ref{alg:Framwork}). Then, the LinUCB algorithm picks the arm with the largest upper confidence bound (step 7 of Algorithm~\ref{alg:Framwork}). %Here $\beta_{t,i}$ is a parameter designed according to the data. 
By following the setup in "optimism in the face of uncertainty linear algorithm" (OFUL) ~\cite{abbasi2011improved}, we set 
\begin{equation}
    \beta_{t,i} = \sqrt{\lambda}S + R\sqrt{2\log K/\delta+d\log\left( 1+ L^2 N_i(t)/(\lambda d) \right)}.
\end{equation}
We define $\omega(N) = \sqrt{\lambda}S + R\sqrt{2\log K/\delta+d\log\left( 1+L^2 N/(\lambda d) \right)}$. It is easy to verify that $\omega(N)$ is a monotonically increasing function over $N \in (0,+\infty)$. 

\begin{algorithm}[htb] 
	\caption{Contextual LinUCB~\cite{li2010contextual}} 
	\label{alg:Framwork} 
	\begin{algorithmic}[1] 
		\REQUIRE ~~\\ 
		regularization $\lambda$, number of arms $K$, number of rounds $T$, bound on context norms $L$, bound on parameter norms $S$.
		\STATE Initialize for every arm $i$, $V_{i} \leftarrow \lambda \mathbf{I}$, $b_{i} \leftarrow \textbf{0}$, $\hat{\theta}_i \leftarrow V_{i}^{-1}b_{i}$.
		\FOR{$t = 1, 2, \dots, T$}
		\STATE observe the context $x_t$.\
		\FOR{$i = 1, 2, \dots, K$}
		\STATE Compute the upper confidence bound: $p_{t,i} \leftarrow x_t^T \hat{\theta}_i  + \beta_{t,i} \sqrt{x_t^T V_{i}^{-1} x_t}$.
		\ENDFOR
		\STATE Pull arm $I_t = \arg \max_i p_{t,i}$.
		\STATE The environment generates reward $r_t$ according to arm $I_t$.
		\STATE The agent receive $r_t$.
		\STATE $V_{i} \leftarrow V_{i}+x_t x_t^T$, $b_i \leftarrow b_{i} + r_t x_t $, $\hat{\theta}_i \leftarrow V_{i}^{-1}b_{i}$.
		\ENDFOR
	\end{algorithmic}
\end{algorithm}

\subsection{White-box Attack} \label{sec:wba}

We first consider the white-box attack scenario, in which the attacker has knowledge of the environment. In particular, in the white-box attack scenario, the attacker knows the coefficient vectors $\theta_i$'s for all $i$. The understanding of this scenario could provide useful insights for the more practical black-box attack scenario to be discussed in Section~\ref{sec:bba}.

The proposed attack strategy works as follows. When the agent chooses arm $K$, the attacker does not attack. When the agent chooses arm $I_t \ne K$, the attacker changes it to arm
\begin{equation}\label{eq:white}
   I_t^0=\left\{
\begin{aligned}
& K &  & \text{with probability} & &\epsilon_t \\
& I_t^\dag &  & \text{with probability} & & 1-\epsilon_t
\end{aligned}
\right.
\end{equation}
where $I_t^\dag = \arg \min_i \langle x_t,\theta_i \rangle$ and
\begin{equation}\label{eq:epsilon}
\epsilon_t = \frac{(1-\alpha) \langle x_t,\theta_K \rangle -\min_i \langle x_t,\theta_i \rangle}{\langle x_t,\theta_K \rangle - \min_i \langle x_t,\theta_i \rangle}.
\end{equation}

We now explain the main idea behind the attack strategy specified in~\eqref{eq:white} and~\eqref{eq:epsilon}. Intuitively speaking, using~\eqref{eq:white} and~\eqref{eq:epsilon}, the attacker can manipulate the agent into learning some particular coefficient vectors. In particular, for arm $K$ (the target arm), the agent obtains the true coefficient vector $\theta_K$. For any arm $i \ne K$, because of the attacks, the agent will obtain an estimate of arm $i$'s coefficient vector to be close to $(1 - \alpha) \theta_K$. This can be seen as follows. Under Assumption \ref{asp:nonworst} and the fact that the mean rewards are positive, we have that $1/2 < \epsilon_t < 1 - \alpha$. Under the white-box attack, when the agent pulls a non-target arm $I_t \ne K$, the mean reward received by the agent should satisfy
\begin{equation}
\begin{aligned}
\mathbb{E}[r_{t,I_t^0}|F_{t-1}, I_t]&= \mathbb{E}[\langle x_t,\theta_{I_t^0} \rangle |F_{t-1}, I_t] \\
&= \epsilon_t\langle x_t,\theta_{K} \rangle + (1-\epsilon_t) \min_{i}\langle x_t,\theta_{i} \rangle \\
& = (1-\alpha)\langle x_t,\theta_{K} \rangle.
\end{aligned}
\end{equation}
Because of the positive mean reward assumption and the fact that $0 < 1-\alpha < 1$, the target arm $K$ appears to be the optimal arm for any context. Hence the proposed white-box attack strategy can easily mislead an agent who employs a contextual bandit algorithm with sublinear pseudo-regret, as shown in the following proposition whose proof can be found in Appendix~\ref{app:white} of the supplementary material.

\begin{proposition}\label{prop:white}
For $\delta > 0$. Suppose the agent employs a contextual bandit algorithm whose cumulative pseudo-regret is upper bounded by $o(T)$ over $T$ with probability $1-\delta$. When using the proposed white-box attack scheme, with probability $1-\delta$, the attacker can force the agent to pull the target arm for $T - o(T)$ times, while the total attack cost $|\mathcal{C}|$ is upper bounded by $o(T)$.
\end{proposition}

Proposition~\ref{prop:white} applies to any contextual bandit algorithm whose cumulative pseudo-regret is upper bounded by $o(T)$. To further illustrate the proposed attack scheme, we now provide a finer analysis the impact of this attack on LinUCB described in Algorithm~\ref{alg:Framwork}. 
We first build a high probability bound for the estimate of parameter at the agent, when the agent computes the estimate of $\theta_i$ by \eqref{ridge} and the attacker employs the white-box attack. Recall that $\omega(N) = \sqrt{\lambda}S + R\sqrt{2\log K/\delta+d\log\left( 1+ L^2 N/(\lambda d)\right)}$. 
\begin{lemma} \label{lem:wb} Under the proposed white-box attack, the estimate of $\theta_i$ for each arm $i$ obtained by LinUCB agent as described in Algorithm~\ref{alg:Framwork} satisfies
\begin{equation}
\begin{aligned}
     |x_t^T\hat{\theta}_{t,i}-x_t^T (1-\alpha) \theta_K| \le  \left( \omega(N_i(t))+LS\sqrt{0.5\log\left(2KT/\delta\right)  }  \right) \Vert x_t \Vert_{V_{t,i}^{-1}} ,
\end{aligned}
\end{equation}
with probability $1-2(K-1)\delta/K$, for all arm $i \ne K$ and all $t \ge 0$. Here, $ \Vert x \Vert_{V} = \sqrt{x^T V x}$ is the weighted norm of vector $x$ for a positive definite matrix $V$. 
\end{lemma}
The proof of  Lemma~\ref{lem:wb} is provided in Appendix~\ref{proofl1} of the supplementary material. Lemma~\ref{lem:wb} shows that, under our white-box attack, the agent's estimate of the parameter of non-target arm, i.e. $\hat{\theta}_i$, will converge to $ (1-\alpha) \theta_K$. Thus, the agent is misled to believe that arm $K$ is the optimal arm for every context in most rounds. The following theorem provides an upper bound of the cumulative cost of the attack.
\begin{theorem} \label{Thmwba}
Define $\gamma = \min_{x \in \mathcal{D}} \langle x,\theta_K \rangle$. Under the same assumptions as in Lemma~\ref{lem:wb}, for any $\delta > 0$ with probability at least $1-2\delta$, for all $T \ge 0$, the attacker can manipulate the LinUCB agent into pulling the target arm in at least $T - |\mathcal{C}|$ rounds, using an attack cost  
\begin{equation}
\begin{aligned}
|\mathcal{C}| \le  \frac{2d(K-1)}{(\alpha \gamma)^2}  \log \left(1+T L^2/(d \lambda)\right) \left( 2\omega(T) 
 +LS\sqrt{0.5\log\left(2KT/\delta\right)  }  \right)^2.
\end{aligned}
\end{equation}

\end{theorem}
The proof of Theorem~\ref{Thmwba} is provided in Appendix~\ref{proofThmwba} of the supplementary material. Theorem~\ref{Thmwba} shows that our white-box attack strategy can successfully force LinUCB agent into pulling the target arm $T-O(\log^2 T)$ times with attack cost scaled only as $O(\log^2 T)$.

\subsection{Black-box Attack}\label{sec:bba}
We now focus on the more practical black-box setting, in which the attacker does not know any of arm's coefficient vector. The attacker knows the value of $\alpha$ (or a lower bound) in which the equation~\eqref{eq:alpha} holds for all $t$. Although the attacker does not know the coefficient vectors for all arms, the attacker can compute an estimate of the unknown parameters by using past observations. On the other hand, there are multiple challenges brought by the estimation errors that need to properly addressed.

The proposed black-box attack strategy works as follows. When the agent chooses arm $K$, the attacker does not attack. When the agent chooses arm $I_t \ne K$, the attacker changes it to arm
\begin{equation}
   I_t^0=\left\{
\begin{aligned}
& K &  & \text{with probability} & &\epsilon_t \\
& I_t^{\dag} &  & \text{with probability} & & 1-\epsilon_t
\end{aligned}
\right.
\end{equation}
where
\begin{equation}
I_t^{\dag} = \arg \min_{i \ne K} \left(\langle x_t, \hat{\theta}_{t,i}^0 \rangle - \beta_{t,i}^0 \Vert x_t \Vert_{(V_{t,i}^0)^{-1}}  \right),\label{eq:lcb}
\end{equation}
and
\begin{equation} 
\begin{aligned}
    \beta_{t,i}^0 = \phi_i \left(\omega(N_i^{\dag}(t))+LS\sqrt{0.5\log\left(2KT/\delta\right)  } \right),
\end{aligned}
\end{equation}
$ \phi_i= 1/\alpha$ when $i \ne K$ and $\phi_K = 2$, and
\begin{equation} \label{eq:epsilon^0}
\begin{aligned}
   \epsilon_t = \text{clip} \left( \frac{1}{2},  \frac{ (1-\alpha) \langle x_t, \hat{\theta}_{t,K}^0 \rangle - \langle x_t,\hat{\theta}_{t,I_t^{\dag}}^0 \rangle}{\langle x_t,\hat{\theta}_{t,K}^0 \rangle -  \langle x_t,\hat{\theta}_{t,I_t^{\dag}}^0 \rangle} , 1-\alpha \right),
\end{aligned}
\end{equation}
with $\text{clip} (a,x,b) = \min(b,\max(x,a))$ where $a \le b$.

For notational convenience, we set $I_t^{\dag} = K$ and $\epsilon_t = 1$ when $I_t = K$. We define that, if $i \ne K$, $\tau_{i}^{\dag}(t):=\{ s:s\le t, I_s^{\dag} = i \}$ and $N_{i}^{\dag}(t)=|\tau_{i}^{\dag}(t)|$ ; $\tau_{K}^{\dag}(t):=\{ s:s\le t \}$ and $N_{K}^{\dag}(t)=|\tau_{K}^{\dag}(t)|$.  
%Here, we use the importance sampling to obtain the $\ell_2$-regularized least-squares estimate of $\theta_i$ with regularization parameter $\lambda > 0$
\begin{equation} \label{ridge^0}
\hat{\theta}_{t,i}^0=\left(V_{t,i}^0  \right)^{-1}\left(\sum_{k \in \tau_{i}^{\dag}(t-1)} w_{k,i}  r_{k,I_k^0} x_k \right) ,
\end{equation}
where $V_{t,i}^0 = \sum_{k \in \tau_{i}^{\dag}(t-1)} x_k x_k^T + \lambda \mathbf{I}$ and 
\begin{equation}
   w_{t,i}=\left\{
\begin{aligned}
& 1/\epsilon_t &  & \text{if} ~ i = I_t^0 = K \\
& 1/(1-\epsilon_t) &  & \text{if} ~  i = I_t^0 = I_t^{\dag} \\
& 0 &  & \text{if} ~  i \ne I_t^0
\end{aligned}
\right..
\end{equation}
Here, $\hat{\theta}_{t,i}^0$ is the estimation of $\theta_i$ by the attacker, while $\hat{\theta}_{t,i}$ in~\eqref{ridge} is the estimation of $\theta_i$ at the agent side. We will show in Lemma \ref{lem:cboftar} and Lemma~\ref{lem:bba} that $\hat{\theta}_{t,i}^0$ will be close to the true value of $\theta_i$ while $\hat{\theta}_{t,i}$ will be close to a sub-optimal value chosen by the attacker. This disparity gives the attacker the advantage and foundation for carrying out the attack.

We now highlight the main idea why our black-box attack strategy works. As discussed in Section~\ref{sec:wba}, if the attacker knows the coefficient vectors of all arms, the proposed white-box attack scheme can mislead the agent to believe that the coefficient vector of every non-target arm is $(1 - \alpha)\theta_K$, hence the agent will think the target arm is optimal. In the black-box setting, the attacker does not know the coefficient vector for any arm. The attacker should estimate the coefficient vector of each arm. Then, the attacker will use the estimated coefficient vector to replace the true coefficient vector in the white-box attack scheme. As the attacker does not know the true values of $\theta_i$'s, we need to design the estimator $\hat{\theta}_{t,i}^0$, the attack choice  $I_t^{\dag}$ and the probability $\epsilon_t$ carefully. In the following, we explain the main ideas behind our design choices. 

Firstly, we explain why we design estimator $\hat{\theta}_{t,i}^0$ using the form~\eqref{ridge^0}, in which the attacker employs the importance sampling to obtain an estimate of $\theta_i$. There are two reasons for this. Firstly, for a successful attack, the number of observation in arm $i\neq K$ will be limited. Hence if the importance sampling is not used, the estimation variance of the mean reward $\langle x,\theta_i \rangle$ at the attacker side for some contexts $x$ may be large.  Secondly, the attacker's action is stochastic when the agent pulls a non-target arm. Thus, the attacker uses the observations at round $t$ when the attacker pulls arm $i$ with certain probability, i.e. when $t \in \tau_{i}^{\dag}$, to estimate $\theta_i$. At the agent side, since the agent's action is deterministic, the agent uses the observations at round $t$ when the agent pulls arm $i$, i.e. when $t \in \tau_{i}$, to estimate $\theta_i$.

Secondly, we explain ideas behind the choice of $I_t^{\dag}$ in~\eqref{eq:lcb}. Under our black-box attack, when the agent pulls a non-target arm $I_t \ne K$, the mean reward received by the agent satisfies
\begin{equation}
\begin{aligned}
\mathbb{E}[r_{t,I_t^0}|F_{t-1}, I_t] & = \mathbb{E}[\langle x_t,\theta_{I_t^0} \rangle |F_{t-1}, I_t] \\
& = \epsilon_t\langle x_t,\theta_{K} \rangle + (1-\epsilon_t)\langle x_t,\theta_{I_t^{\dag}} \rangle.
\end{aligned}
\end{equation}
In our white-box attack scheme, $I_t^\dag$ is the worst arm at context $x_t$. In the black-box setting, the attack does not know a prior which arm is the worst. In the proposed black-box attack scheme, as indicated in~\eqref{eq:lcb}, we use the lower confidence bound (LCB) method to explore the worst arm and $I_t^\dag$ is the arm whose lower confidence bound is the smallest. 

Finally, we provide reasons why we choose $\epsilon_t$ using~\eqref{eq:epsilon^0}. In our white-box attack scheme, we have that $1/2 < \epsilon_t < 1 - \alpha$. Thus, in our black-box attack scheme, we limit the choice of $\epsilon_t$ to $[1/2, 1 - \alpha]$. Furthermore, in~\eqref{eq:epsilon} used for the white-box attack, $\epsilon_t$ is computed by the true mean reward. Now, in the black-box attack, as the attacker does not the true coefficient vector, the attacker use the estimation of $\theta$ to compute the second term in the clip function in~\eqref{eq:epsilon^0}.

In summary, intuitively speaking, our design of $\hat{\theta}_{t,i}^0$, $I_t^{\dag}$ and $\epsilon_t$ can ensure that the attacker's estimation $\hat{\theta}_{t,i}^0$ will be close to $\theta_i$, while the agent's estimation $\hat{\theta}_{t,i}$ will be close to $(1 - \alpha)\theta_K$. In the following, we make these statements precise, and formally analyze the performance of the proposed black-box attack scheme.

First, we analyze the estimation $\hat{\theta}_{t,i}^0$ at the attacker side. We establish a confidence ellipsoid of $\langle x_t, \hat{\theta}_{t,i}^0 \rangle $ at the attacker.
\begin{lemma} \label{lem:cboftar}
Assume the attacker performs the proposed black-box action poisoning attack. With probability $1-2\delta$, we have 
\begin{equation} 
\begin{aligned}
    |x_t^T\hat{\theta}_{t,i}^0-x_t^T\theta_i| \le  \beta_{t,i}^0 \Vert x_t \Vert_{\left(V_{t,i}^0  \right)^{-1}}.
\end{aligned}
\end{equation}
holds for all arm $i$ and all $t \ge 0$ simultaneously. 
\end{lemma}
 Lemma~\ref{lem:cboftar} shows that $\hat{\theta}_{i}^0$ lies  in an ellipsoid with center at $\theta_i$ with high probability, which implies that the attacker has good estimate of each arm.

We then analyze the estimation $\hat{\theta}_{t,i}$ at the agent side. The following lemma provides an upper bound on the absolute difference between $\mathbb{E}[r_{t,I_t^0}|F_{t-1}, I_t]$ and $(1-\alpha)\langle x_t,\theta_{K} \rangle$.

\begin{lemma} \label{lem:difbtw}
Under the black-box attack, with probability $1-2\delta$, the estimate obtained by an LinUCB agent satisfies
\begin{equation}
\begin{aligned}
& \left|\mathbb{E}[r_{t,I_t^0}|F_{t-1}, I_t] - (1-\alpha) \langle  x_t,  \theta_K \rangle \right|\\
\le & (1-\alpha)\beta_{t,K}^0 \Vert x_t \Vert_{\left(V_{t,K}^0  \right)^{-1}}  + (1+\alpha) \beta_{t,I_t^{\dag}}^0 \Vert x_t \Vert_{\left(V_{t,I_t^{\dag}}^0  \right)^{-1}} .
\end{aligned}\nonumber
\end{equation}
 simultaneously for all $t \ge 0$ when $I_t \ne K$.
\end{lemma}

The bound in Lemma~\ref{lem:difbtw} consists of the confidence ellipsoid of the estimate of arm $I_t^{\dag}$ and that of arm $K$. As mentioned above, for a successful attack, the number of observations on arm $I_t^{\dag}$ will be limited. Thus, in our proposed algorithm, the attacker use the importance sampling to obtain the estimate of $\theta_i$, which will increases the number of observations that can be used to estimate the coefficient vector of arm $I_t^{\dag}$. Using Lemma~\ref{lem:difbtw}, we have the following lemma regarding the estimation $\hat{\theta}_{t,i}$ at the agent side.

\begin{lemma} \label{lem:bba}
Consider the same assumption as in Lemma~\ref{lem:cboftar}. With a probability at least $1-\frac{(3K-1)\delta}{K}$, the estimate $\hat{\theta}_{t,i}$ obtained by the LinUCB agent will satisfy
\begin{equation}
\begin{aligned}
    & |x_t^T\hat{\theta}_{t,i}-x_t^T (1-\alpha) \theta_K| \\
 \le & \left( 1+ 4d /\alpha\sqrt{K\log \left(1+t L^2/(d \lambda)\right)} \right)  \left(  \omega(N_i(t))+  LS\sqrt{0.5\log\left(2KT/\delta\right)}\right) \Vert x_t \Vert_{V_{t,i}^{-1}} ,
\end{aligned}
\end{equation}
simultaneously for all arm $i \ne K$ and all $t \ge 0$. 
\end{lemma}

Lemma~\ref{lem:bba} shows that, under the proposed black-box attack scheme, the agent's estimate of the parameter of non-target arm, i.e. $\hat{\theta}_i$, will converge to $ (1-\alpha) \theta_K$. As the result, the agent will believe that the target arm $K$ is the optimal arm for any context in most rounds. Using these supporting lemmas, we can then analyze the performance of the proposed black-box attack strategy.

\begin{theorem} \label{thm:bbacost}
Under the same assumptions as in Lemma~\ref{lem:bba}, with probability at least $1-3\delta$, for all $T \ge 0$, the attacker can manipulate a LinUCB agent into pulling the target arm in at least $T - |\mathcal{C}|$ rounds, using an attack cost  
\begin{equation}
\begin{split}
& |\mathcal{C}| \le  \frac{2d(K-1)}{(\alpha \gamma)^2} \left( 2+ \frac{4d }{\alpha}\sqrt{K\log \left(1+\frac{T L^2}{d \lambda}\right)} \right)^2  \log \left(1+T L^2/(d \lambda)\right) \left(  \omega(T)+  LS\sqrt{0.5\log\left(2KT/\delta\right)}\right)^2.
\end{split}\nonumber
\end{equation}
\end{theorem}

Theorem~\ref{thm:bbacost} shows that our black-box attack strategy can manipulate a LinUCB agent into pulling a target arm $T-O(\log^3 T)$ times with attack cost scaling as $O(\log^3 T)$. Compared with the result for the white-box attack, the black-box attack only brings an additional $\log T$ factor.

\section{Numerical Experiments}

In this section, we empirically evaluate the performance of the proposed action poisoning attack schemes on three contextual bandit algorithms: LinUCB~\cite{abbasi2011improved}, LinTS~\cite{pmlr-v28-agrawal13}, and $\epsilon$-Greedy. We run the experiments on three datasets:

 \textbf{Synthetic data:} The dimension of contexts and the coefficient vectors is $d=6$. We set the first entry of every context and coefficient vector to $1$. The other entries of every context and coefficient vector are uniformly drawn from $(-\frac{1}{\sqrt{d-1}}, \frac{1}{\sqrt{d-1}})$. Thus, for all round $t$ and arm $i$, $\Vert x_t \Vert_2 \le \sqrt{2}$, $\Vert \theta_i \Vert_2 \le \sqrt{2}$ and mean rewards $\langle x_t, \theta_i \rangle > 0$. The reward noise $\eta_t$ is drawn from a Gaussian distribution $\mathcal{N}(0,0.01)$.
   
   \textbf{Jester dataset~\cite{goldberg2001eigentaste}:} Jester contains 4.1 million ratings of jokes in which the rating values scale from $-10.00$ to $+10.00$. We normalize the rating to $[0,1]$. The dataset includes 100 jokes and the ratings were collected from 73,421 users between April 1999 - May 2003. We consider a subset of 10 jokes and 38432 users. Every jokes are rated by each user. We perform a low-rank matrix factorization ($d=6$) on the ratings data and obtain the features for both users and jokes. At each round, the environment randomly select a user as the context and the reward noise is drawn from a Gaussian distribution $\mathcal{N}(0,0.01)$.

 \textbf{MovieLens 25M dataset:~\cite{harper2015movielens}} MovieLens 25M dataset contains 25 million 5-star ratings of 62,000 movies by 162,000 users. The preprocessing of this data is almost the same as the Jester dataset, except that we consider a subset of 10 movies and 7344 users. At each round, the environment randomly select a user as the context and the reward noise is drawn from a Gaussian distribution $\mathcal{N}(0,0.01)$.

We set $\delta = 0.1$ and $\lambda = 2$. For all the experiments, we set the total number of rounds $T = 10^6$ and the number of arms $K = 10$. We independently run ten repeated experiments. Results reported are averaged over the ten experiments.

\begin{figure*}[t]
\centering
\includegraphics[width=0.9\textwidth]{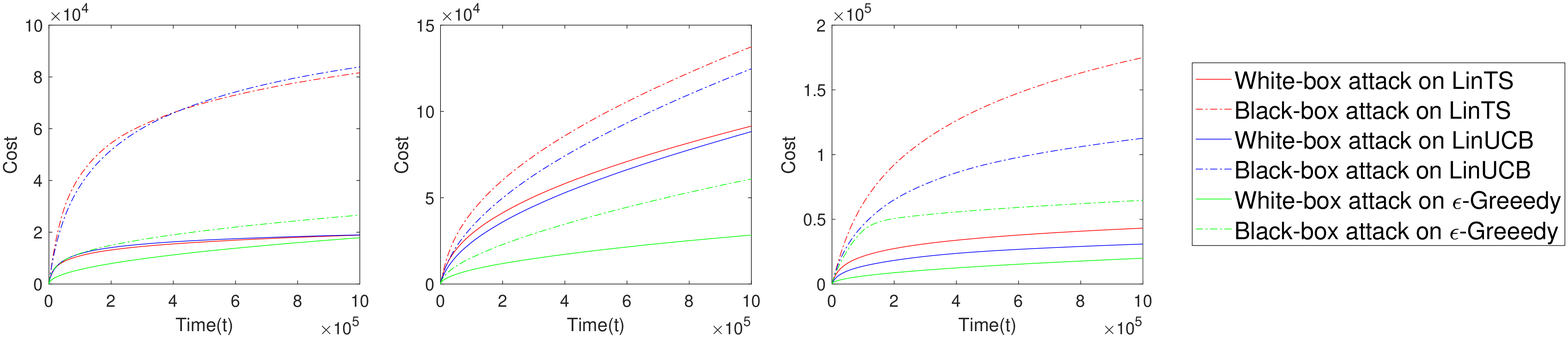}  %Reduce the figure size so that it is slightly narrower than the column.
\caption{The cumulative cost of the attacks for the synthetic (Left), Jester (Center) and MovieLens (Right) datasets.}
\label{fig1}
\end{figure*}

The results are shown in Table~\ref{table1} and Figure~\ref{fig1}. These experiments show that the action poisoning attacks can force the three agents to pull the target arm very frequently, while the agents rarely pull the target arm under no attack. Under the attacks, the true regret of the agent becomes linear as the target arm is not optimal for most context. Table~\ref{table1} show the number of rounds the agent pulls the target arm among $T=10^6$ total rounds. In the synthetic dataset, under the proposed white-box attacks, the target arm is pulled more than $98.1\%$ of the times by the three agent (see Table~\ref{table1}). The target arm is pulled more than $91.6\%$ of the times in the worst case (the black-box attacks on LinUCB). Fig~\ref{fig1} shows the cumulative cost of the attacks on three agents for the three datasets. The results show that the attack cost $|\mathcal{C}|$ of every attack scheme on every agent for every dataset scales sublinearly, which exposes a significant security threat of the action poisoning attacks on linear contextual bandits. 

\begin{table*}[t]
\centering

\begin{tabular}{|l|l|l|l|}
\hline
     & Synthetic & Jester & MovieLens \\
\hline
    $\epsilon$-Greeedy without attacks & ~~~~2124.6 & ~~~~5908.7 & ~~~~3273.5 \\
    White-box attack on $\epsilon$-Greeedy & 982122.5 & 971650.9 & 980065.6 \\
    Black-box attack on $\epsilon$-Greeedy & 973378.5 & 939090.2 & 935293.8 \\
    LinUCB without attacks & ~~~~8680.9 & ~~16927.2 & ~~13303.4 \\
    White-box attack on LinUCB & 981018.7 & 911676.9 & 969118.6 \\
    Black-box attack on LinUCB & 916140.8 & 875284.7 & 887373.1 \\
    LinTS without attacks & ~~~~5046.9 & ~~18038.0 & ~~~~9759.0 \\
    White-box attack on LinTS & 981112.8 & 908488.3 & 956821.1 \\
    Black-box attack on LinTS & 918403.8 & 862556.8 & 825034.8 \\
     \hline
\end{tabular}

\caption{Average number of rounds when the agent pulls the target arm over $T=10^6$ rounds.}
\label{table1}
\end{table*}

\section{Conclusion}
In this paper, we have proposed a new class of attacks on linear contextual bandits: action poisoning attacks. %We compared the action poisoning attacks with the reward action attacks and the context action attacks. 
We have shown that our white-box attack strategy is able to force any linear contextual bandit agent, whose regret scales sublinearly with the total number of rounds, into pulling a target arm chosen by the attacker. In addition, we have shown that our white-box attack strategy can force LinUCB agent into pulling a target arm $T-O(\log^2 T)$ times with attack cost scaled as $O(\log^2 T)$. We have further shown that the proposed black-box attack strategy can force LinUCB agent into pulling a target arm $T-O(\log^3 T)$ times with attack cost scaled as $O(\log^3 T)$. Our results expose a significant security threat to contextual bandit algorithms. In the future, we will investigate the defense strategy to mitigate the effects of this attack.

\bibliographystyle{IEEEbib}
\bibliography{mybib}

\newpage
\appendix

\section{Proof of Proposition~\ref{prop:white}}\label{app:white}

	When the agent pulls a non-target arm $I_t  \ne K$, the mean reward received by the agent should satisfy $\mathbb{E}[r_{t,I_t^0}|F_{t-1}, I_t] = (1-\alpha)\langle x_t,\theta_{K} \rangle$. In the observation of the agent, the target arm becomes optimal and the non-target arms are associated with the coefficient vector $(1-\alpha)\theta_{K}$. In addition, the cumulative pseudo-regret should satisfy 
	$\bar{R}_T=\sum_{t=1}^T \mathbbm{1}_{\{ I_t \ne K \}} \alpha \langle x_t,\theta_{K} \rangle $. If $\bar{R}_T$ is upper bounded by $o(T)$,  $\sum_{t=1}^T \mathbbm{1}_{\{ I_t \ne K \}}$ is also upper bounded by $o(T)$.

\section{Proof of Lemma~\ref{lem:wb}} \label{proofl1}
If the agent computes an estimate of $\theta_i$ by \eqref{ridge} and $V_{t,i} = \left(\sum_{k \in \tau_i(t-1)} x_{k} x_{k}^T + \lambda \mathbf{I}\right)$, we have 
\begin{equation}
\begin{aligned}
    &x_t^T\hat{\theta}_{t,i}-x_t^T (1-\alpha) \theta_K\\
    =&x_t^T V_{t,i}^{-1} \left(\sum_{k \in \tau_i(t-1)} r_{t,I_k^0} x_{k}\right)-x_t^T V_{t,i}^{-1} V_{t,i} (1-\alpha) \theta_K \\
    =&x_t^T V_{t,i}^{-1} \left(\sum_{k \in \tau_i(t-1)} x_{k}\left(r_{t,I_k^0}-(1-\alpha) x_{k}^T\theta_K\right)\right) -\lambda x_t^T V_{t,i}^{-1}  (1-\alpha) \theta_K\\
    =&\sum_{k  \in \tau_i(t-1)} x_t^T V_{t,i}^{-1} x_{k}\left( x_{k}^T\theta_{I_k^0}+\eta_k-(1-\alpha) x_{k}^T\theta_K\right)  - \lambda x_t^T V_{t,i}^{-1} (1-\alpha) \theta_K,
\end{aligned}
\end{equation}
and by triangle inequality,
\begin{equation} 
\begin{aligned}
    &|x_t^T\hat{\theta}_{t,i}-x_t^T (1-\alpha) \theta_K|\\
    \le&\left|\sum_{k  \in \tau_i(t-1)} x_t^T V_{t,i}^{-1} x_{k}\left( x_{k}^T\theta_{I_k^0}-(1-\alpha) x_{k}^T\theta_K\right)\right|  + \left|\sum_{k  \in \tau_i(t-1)} x_t^T V_{t,i}^{-1} x_{k} \eta_k \right|+ \left|\lambda x_t^T V_{t,i}^{-1} (1-\alpha) \theta_K \right|.
\end{aligned}
\end{equation}

In our model, the mean reward is bounded by $ 0<\langle x_t,\theta_i \rangle \le \Vert x_t \Vert_2^2 \Vert \theta_i \Vert_2^2 = LS$. Since the mean rewards are bounded and the rewards are generated independently, we have $0 \le \left| x_{k}^T\theta_{I_k^0} - (1-\alpha) x_{k}^T\theta_K \right| \le LS$ and 
$\mathbb{E}[ x_{k}^T\theta_{I_k^0} | F_{k-1} ]= (1-\alpha) x_{k}^T\theta_K $. Thus, $\left\{ x_t^T V_{t,i}^{-1} x_{k}\left( x_{k}^T\theta_{I_k^0}-(1-\alpha) x_{k}^T\theta_K\right)\right\}_{k \in \tau_i(t-1)}$ is a bounded martingale difference sequence w.r.t the filtration $\{F_k\}_{k \in \tau_i(t-1)}$. 

Then, by Azuma's inequality,
\begin{equation}\label{eq:Azuma}
\begin{aligned}
    &\mathbb{P} \left( \left|\sum_{k  \in \tau_i(t-1)} x_t^T V_{t,i}^{-1} x_{k}\left( x_{k}^T\theta_{I_k^0}-(1-\alpha) x_{k}^T\theta_K\right)\right| \ge B \right)\\
    \le& 2 \exp \left(\frac{-2 B^2 }{ \sum_{k  \in \tau_i(t-1)}(x_t^T V_{t,i}^{-1} x_{k} LS)^2} \right)\\
    =& P_{i,t},
\end{aligned}
\end{equation}
where $B$ represents confidence bound. In order to ensure the confidence bounds hold for all arms and all round $t$ simultaneously, we set $P_{i,t}=\frac{\delta}{KT}$ so
\begin{equation}
\begin{aligned}
B & = LS\sqrt{\frac{1}{2}\log\left(\frac{2KT}{\delta}\right) \sum_{k  \in \tau_i(t-1)}(x_t^T V_{t,i}^{-1} x_{k})^2 } \\
& \le   LS\sqrt{\frac{1}{2}\log\left(\frac{2KT}{\delta}\right)  } \Vert x_t \Vert_{V_{t,i}^{-1}},
\end{aligned}
\end{equation}
where the  last inequality is obtained from the fact that 
\begin{equation} \label{fact1}
\begin{aligned}
\Vert x_t \Vert_{V_{t,i}^{-1}}^2 &= x_t^T V_{t,i}^{-1} \left(\sum_{k \in \tau_i(t-1)} x_{k} x_{k}^T + \lambda \mathbf{I} \right) V_{t,i}^{-1} x_t \\
& \ge x_t^T V_{t,i}^{-1} \left(\sum_{k \in \tau_i(t-1)} x_{k} x_{k}^T \right) V_{t,i}^{-1} x_t \\
& = \sum_{k  \in \tau_i(t-1)}(x_t^T V_{t,i}^{-1} x_{k})^2.
\end{aligned}
\end{equation}

In other words, with probability $1-\delta$, we have
\begin{equation}
\begin{aligned}
     & \left|\sum_{k  \in \tau_i(t-1)} x_t^T V_{t,i}^{-1} x_{k}\left( x_{k}^T\theta_{I_k^0}-(1-\alpha) x_{k}^T\theta_K\right)\right| \\
      \le & LS\sqrt{\frac{1}{2}\log\left(\frac{2KT}{\delta}\right)  } \Vert x_t \Vert_{V_{t,i}^{-1}},
\end{aligned}
\end{equation}
for all arms and all $t$. 

Note that $V_{t,i} = \sum_{k \in \tau_i(t-1)} x_{k} x_{k}^T + \lambda \mathbf{I}$ is positive definite. We define $\langle x, y \rangle_V = x^T V y $ as the weighted inner-product. According to Cauchy-Schwarz inequality, we have
\begin{equation}
\begin{aligned}
\left|\sum_{k  \in \tau_i(t-1)} x_t^T V_{t,i}^{-1} x_{k} \eta_k \right|  
\le \left\Vert x_t \right\Vert_{V_{t,i}^{-1}} \left\Vert \sum_{k  \in \tau_i(t-1)}  x_{k} \eta_k \right\Vert_{V_{t,i}^{-1}}.
\end{aligned}
\end{equation}

Assume that $\lambda \ge L$. From Theorem 1 and Lemma 11 in~\cite{abbasi2011improved}, we know that for any $\delta > 0$, with probability at least $1-\delta$
\begin{equation} \label{eq:lin2011thm1}
\begin{aligned}
 &\left\Vert \sum_{k  \in \tau_i(t-1)}  x_{k} \eta_k \right\Vert_{V_{t,i}^{-1}}^2 \\
\le & 2R^2 \log\left(\frac{K\det(V_{t,i})^{1/2}\det(\lambda \mathbf{I})^{-1/2}}{\delta}\right) \\
\le & R\sqrt{2\log\frac{K}{\delta}+d\log\left( 1+ \frac{L^2 N_i(t)}{\lambda d} \right)},
\end{aligned}
\end{equation}
for all arms and all $t>0$,

For the third part of the right hand side of~\eqref{eq:wbdivide},
\begin{equation}
\begin{aligned}
|\lambda x_t^T V_{t,i}^{-1}  (1-\alpha) \theta_K| &  \le \left\Vert (1-\alpha) \lambda \theta_K \right\Vert_{V_{t,i}^{-1}}  \left\Vert x_t \right\Vert_{V_{t,i}^{-1}} .
\end{aligned}
\end{equation}

Since $V_{t,i} \succeq \lambda \mathbf{I}$, the maximum eigenvalue of $V_{t,i}^{-1}$ is smaller or equal to $1/\lambda$. Thus, $\left\Vert (1-\alpha) \lambda \theta_K \right\Vert_{V_{t,i}^{-1}}^2 \le \frac{1}{\lambda} \left\Vert (1-\alpha) \lambda \theta_K \right\Vert_2^2 \le (1-\alpha)^2\lambda S^2$.

In summary, 
\begin{equation} \label{eq:wbdivide}
\begin{aligned}
    &|x_t^T\hat{\theta}_{t,i}-x_t^T (1-\alpha) \theta_K|\\
    \le& \left( (1-\alpha) \sqrt{\lambda} S+LS\sqrt{\frac{1}{2}\log\left(\frac{2KT}{\delta}\right)  }  + R\sqrt{2\log\frac{K}{\delta}+d\log\left( 1+ \frac{L^2 N_i(t)}{\lambda d} \right)} \right) \Vert x_t \Vert_{V_{t,i}^{-1}}.
\end{aligned}
\end{equation}

\section{Proof of Theorem 1} \label{proofThmwba}
For round $t$ and context $x_t$, if LinUCB pulls arm $i \ne K$, we have
\begin{equation}
\begin{aligned}
x_t^T \hat{\theta}_{t,K} + \beta_{t,K} \sqrt{x_t^T V_{t,K}^{-1} x_t} \le x_t^T \hat{\theta}_{t,i} + \beta_{t,i} \sqrt{x_t^T V_{t,i}^{-1} x_t}.\nonumber
\end{aligned}
\end{equation}
Recall $\beta_{t,i} = \sqrt{\lambda}S + R\sqrt{2\log\frac{K}{\delta}+d\log\left( 1+ \frac{L^2 N_i(t)}{\lambda d} \right)}$.

Since the attacker does not attack the target arm, the confidence bound of arm $K$ does not change and $x_t^T \theta_K \le x_t^T \hat{\theta}_{t,K} + \beta_{t,K} \sqrt{x_t^T V_{t,K}^{-1} x_t} $ holds with probability $1-\frac{\delta}{K}$.

Thus, by Lemma 1,
\begin{equation}
\begin{aligned}
x_t^T \theta_K \le & x_t^T \hat{\theta}_{t,i} + \beta_{t,i} \sqrt{x_t^T V_{t,i}^{-1} x_t} \\
\le&
x_t^T (1-\alpha) \theta_K + \left( LS\sqrt{\frac{1}{2}\log\left(\frac{2KT}{\delta}\right)  }   \omega(N_i(t)) + \beta_{t,i} \right)  \Vert x_t \Vert_{V_{t,i}^{-1}}.
\end{aligned}
\end{equation}
By multiplying both sides $\mathbbm{1}_{\{ I_t = i \}}$ and summing over rounds, we have
\begin{equation}
\begin{aligned}
 & \sum_{k=1}^t \mathbbm{1}_{\{ I_k = i \}} \alpha x_k^T \theta_K \\
\le  & \sum_{k=1}^t \mathbbm{1}_{\{ I_k = i \}} \left(\beta_{k,i}+ \sqrt{\lambda}S+LS\sqrt{\frac{1}{2}\log\left(\frac{2KT}{\delta}\right)  }   + R\sqrt{2\log\frac{K}{\delta}+d\log\left( 1+ \frac{L^2 N_i(k)}{\lambda d} \right)} \right)  \Vert x_k \Vert_{V_{k,i}^{-1}}.
\end{aligned}
\end{equation}

Here, we use Lemma 11 from \cite{abbasi2011improved} and obtain 
\begin{equation}
\begin{aligned}
\sum_{k=1}^t \mathbbm{1}_{\{ I_k = i \}}  \Vert x_k \Vert_{V_{k,i}^{-1}}^2 & \le 2d\log (1+\frac{N_i(t) L^2}{d \lambda}) \le 2d\log \left(1+\frac{t L^2}{d \lambda}\right).
\end{aligned}
\end{equation}

According to $ \sum_{k=1}^t \mathbbm{1}_{\{ I_k = i \}} \Vert x_k \Vert_{V_{k,i}^{-1}} \le \sqrt{N_i(t) \sum_{k=1}^t \mathbbm{1}_{\{ I_k = i \}}  \Vert x_k \Vert_{V_{k,i}^{-1}}^2}$, we have
\begin{equation}
\begin{aligned}
\sum_{k=1}^t \mathbbm{1}_{\{ I_k = i \}}  \Vert x_k \Vert_{V_{k,i}^{-1}}^2 \le \sqrt{N_i(t)2d\log \left(1+\frac{t L^2}{d \lambda}\right)}.
\end{aligned}
\end{equation}

Thus, we have
\begin{equation}
\begin{aligned}
 & \sum_{k=1}^t \mathbbm{1}_{\{ I_k = i \}} \alpha x_k^T \theta_K \\
\le  & \sqrt{N_i(t)2d\log \left(1+\frac{t L^2}{d \lambda}\right)} \left( LS\sqrt{\frac{1}{2}\log\left(\frac{2KT}{\delta}\right)  }  + 2\sqrt{\lambda}S+  2R\sqrt{2\log\frac{K}{\delta}+d\log\left( 1+ \frac{t L^2}{\lambda d} \right)}\right),
\end{aligned}
\end{equation}
and
\begin{equation}
\begin{aligned}
 & N_i(t) = \sum_{k=1}^t \mathbbm{1}_{\{ I_k = i \}}  \\
\le  & \frac{2d}{(\alpha \gamma)^2} \log\left (1+\frac{t L^2}{d \lambda}\right) \left( 2\sqrt{\lambda}S+LS\sqrt{\frac{1}{2}\log\left(\frac{2KT}{\delta}\right)  }   + 2R\sqrt{2\log\frac{K}{\delta}+d\log\left( 1+ \frac{t L^2}{\lambda d} \right)}\right)^2,
\end{aligned}
\end{equation}
where $\gamma = \min_{x \in \mathcal{D}} \langle x,\theta_K \rangle$.

\section{Proof of Lemma~\ref{lem:cboftar}} \label{prfcboftar}

Since the estimate of $\theta_i$ obtained by the agent satisfies
\begin{equation} 
\hat{\theta}_{t,i}^0=\left(V_{t,i}^0  \right)^{-1}\left(\sum_{k \in \tau_{i}^{\dag}(t-1)} w_{k,i} r_{k,I_k^0} x_k \right),
\end{equation}
we have 
\begin{equation} \label{eq:ISdivide}
\begin{aligned}
 & x_t^T\hat{\theta}_{t,i}^0-x_t^T\theta_i \\
 = & x_t^T\left(V_{t,i}^0  \right)^{-1}\left(\sum_{k \in \tau_{i}^{\dag}(t-1)} w_{k,i} r_{k,I_k^0} x_k \right) - x_t^T \left(V_{t,i}^0  \right)^{-1} V_{t,i}^0\theta_i\\
 = & x_t^T\left(V_{t,i}^0  \right)^{-1}\left(\sum_{k \in \tau_{i}^{\dag}(t-1)} (w_{k,i} r_{k,I_k^0}- x_k^T \theta_i) x_k \right) - \lambda x_t^T \left(V_{t,i}^0  \right)^{-1} \theta_i \\
 = & x_t^T\left(V_{t,i}^0  \right)^{-1}\left(\sum_{k \in \tau_{i}^{\dag}(t-1)} (w_{k,i} x_k^T \theta_{I_k^0} - x_k^T \theta_i) x_k \right)  +  x_t^T\left(V_{t,i}^0  \right)^{-1}\left(\sum_{k \in \tau_{i}^{\dag}(t-1)} w_{k,i} \eta_k \right)- \lambda x_t^T \left(V_{t,i}^0  \right)^{-1} \theta_i.\nonumber
\end{aligned}
\end{equation}

We have $0 \le \left|w_{k,i} x_k^T \theta_{I_k^0} - x_k^T \theta_i \right| \le w_{k,i}LS$ and 
$\mathbb{E}[ w_{k,i} x_{k}^T\theta_{I_k^0} | F_{k-1} ]= x_k^T \theta_i $. In addition, by the definition of  $ w_{k,i} $, we have that $ w_{k,i} \le 1/\alpha$ if $i \ne K$, and $ w_{k,i} \le 2$ if $i = K$. 
Thus, $\left\{ x_t^T \left(V_{t,i}^0  \right)^{-1}\left(\sum_{k \in \tau_{i}^{\dag}(t-1)} (w_{k,i} x_k^T \theta_{I_k^0} - x_k^T \theta_i) x_k \right)\right\}_{k \in \tau_i(t-1)}$ is also a bounded martingale difference sequence w.r.t the filtration $\{F_k\}_{k \in \tau_i(t-1)}$. By following the steps in Section~\ref{proofl1}, we have, with probability $1-\frac{K-1}{K}\delta$, for any arm $i \ne K$ and any round $t$,
\begin{equation}
\begin{aligned}
   & \left|x_t^T\left(V_{t,i}^0  \right)^{-1}\left(\sum_{k \in \tau_{i}^{\dag}(t-1)} (w_{k,i} x_k^T \theta_{I_k^0} - x_k^T \theta_i) x_k \right)\right| \\
    \le & \frac{LS}{\alpha}\sqrt{\frac{1}{2}\log\left(\frac{2KT}{\delta}\right)  } \Vert x_t \Vert_{\left(V_{t,i}^0  \right)^{-1}},\nonumber
\end{aligned}
\end{equation}
and with probability $1-\frac{1}{K}\delta$, for arm $K$ and any round $t$,
\begin{equation}
\begin{aligned}
   & \left|x_t^T\left(V_{t,K}^0  \right)^{-1}\left(\sum_{k \in \tau_{i}^{\dag}(t-1)} (w_{k,K} x_k^T \theta_{I_k^0} - x_k^T \theta_K) x_k \right)\right| \\
   \le & 2LS\sqrt{\frac{1}{2}\log\left(\frac{2KT}{\delta}\right)  } \Vert x_t \Vert_{\left(V_{t,K}^0  \right)^{-1}}.\nonumber
\end{aligned}
\end{equation}

The confidence bound of the second item of the right side hand of~\eqref{eq:ISdivide} can be obtained from~\eqref{eq:lin2011thm1}. With probability, $1-\frac{K-1}{K}\delta$, for any arm $i \ne K$ and any round $t$, 
\begin{equation}
\begin{aligned}
& \left|  x_t^T\left(V_{t,i}^0  \right)^{-1}\left(\sum_{k \in \tau_{i}^{\dag}(t-1)} w_{k,i} \eta_k \right) \right|\\
\le & \frac{R}{\alpha}\sqrt{2\log\frac{K}{\delta}+d\log\left( 1+ \frac{L^2 N_i^{\dag}(t)}{\lambda d} \right)} \Vert x_t \Vert_{\left(V_{t,i}^0  \right)^{-1}}.
\end{aligned}
\end{equation}
With probability, $1-\frac{1}{K}\delta$, for arm $K$ and any round $t$, 
\begin{equation} 
\begin{aligned}
& \left|  x_t^T\left(V_{t,i}^0  \right)^{-1}\left(\sum_{k \in \tau_{i}^{\dag}(t-1)} w_{k,i} \eta_k \right) \right|\\
\le & 2R\sqrt{2\log\frac{K}{\delta}+d\log\left( 1+ \frac{L^2 N_K^{\dag}(t)}{\lambda d} \right)} \Vert x_t \Vert_{\left(V_{t,K}^0  \right)^{-1}}.
\end{aligned}
\end{equation}
In summary, 
\begin{equation} 
\begin{aligned}
    & |x_t^T\hat{\theta}_{t,i}^0-x_t^T\theta_i| \\
    \le & \phi_i \left( \sqrt{\lambda} S+LS\sqrt{\frac{1}{2}\log\left(\frac{2KT}{\delta}\right)  } +  R\sqrt{2\log\frac{K}{\delta}+d\log\left( 1+ \frac{L^2 N_i^{\dag}(t)}{\lambda d} \right)} \right) \Vert x_t \Vert_{\left(V_{t,K}^0  \right)^{-1}},
\end{aligned}
\end{equation}
where $\phi_i = 1/\alpha$ when $i \ne K$ and $\phi_K = 2$.

\section{Proof of Lemma~\ref{lem:difbtw}}
Recall the definition of $\epsilon_t$:
\begin{equation}
\begin{aligned}
   \epsilon_t = \text{clip} \left( \frac{1}{2},  \frac{ (1-\alpha) \langle x_t, \hat{\theta}_{t,K}^0 \rangle - \langle x_t,\hat{\theta}_{t,I_t^{\dag}}^0 \rangle}{\langle x_t,\hat{\theta}_{t,K}^0 \rangle -  \langle x_t,\hat{\theta}_{t,I_t^{\dag}}^0 \rangle} , 1-\alpha \right),
\end{aligned}
\end{equation}
and the definition of $I_t^{\dag}$:
\begin{equation}
I_t^{\dag} = \arg \min_{i \ne K} \left(\langle x_t, \hat{\theta}_{t,i}^0 \rangle - \beta_{t,i}^0 \Vert x_t \Vert_{(V_{t,i}^0)^{-1}}  \right).
\end{equation}
By Lemma~\ref{lem:cboftar},  $\langle x_t, \hat{\theta}_{t,I_t^{\dag}}^0 \rangle - \beta_{t,I_t^{\dag}}^0 \Vert x_t \Vert_{(V_{t,I_t^{\dag}}^0)^{-1}} \le \min_i  \langle x_t, \theta_i \rangle$ with probability $1-2\delta$.

Because $\epsilon_t$ is bounded by $[1/2, 1-\alpha]$, we can analyze $\mathbb{E}[r_{t,I_t^0}|F_{t-1}, I_t]$ in four cases.

\textbf{Case 1}: when $\langle x_t,\hat{\theta}_{t,K}^0 \rangle <  \langle x_t,\hat{\theta}_{t,I_t^{\dag}}^0 \rangle$ and $\epsilon_t = 1 - \alpha$, we have
\begin{equation}
    \mathbb{E}[r_{t,I_t^0}|F_{t-1}, I_t] = (1-\alpha)\langle x_t, \theta_K \rangle + \alpha\langle x_t,\theta_{I_t^{\dag}} \rangle.
\end{equation}
Then, by Lemma~\ref{lem:cboftar},
\begin{equation} \label{eq:case1}
\begin{aligned}
 &(1-\alpha)  x_t^T\theta_K + \alpha x_t^T \theta_{I_t^{\dag}}- (1-\alpha) x_t^T\theta_{K} \\
 \le &  (1-\alpha)\left(x_t^T\hat{\theta}_{t,K}^0 + \beta_{t,K}^0 \Vert x_t \Vert_{\left(V_{t,K}^0  \right)^{-1}} \right) + \alpha \left( x_t^T \hat{\theta}_{t,I_t^{\dag}}^0 +  \beta_{t,I_t^{\dag}}^0 \Vert x_t \Vert_{\left(V_{t,I_t^{\dag}}^0  \right)^{-1}}  \right)  - (1-\alpha) x_t^T\theta_{K} \\
\le & x_t^T \hat{\theta}_{t,I_t^{\dag}}^0 - (1-\alpha) x_t^T\theta_{K} +   (1-\alpha) \beta_{t,K}^0 \Vert x_t \Vert_{\left(V_{t,K}^0  \right)^{-1}}  +  \alpha  \beta_{t,I_t^{\dag}}^0 \Vert x_t \Vert_{\left(V_{t,I_t^{\dag}}^0  \right)^{-1}}   \\
  \le  & (1-\alpha) \beta_{t,K}^0 \Vert x_t \Vert_{\left(V_{t,K}^0  \right)^{-1}}  +  (1+\alpha)   \beta_{t,I_t^{\dag}}^0 \Vert x_t \Vert_{\left(V_{t,I_t^{\dag}}^0  \right)^{-1}},
\end{aligned}
\end{equation}
where the last inequality is obtained by $ x_t^T \hat{\theta}_{t,I_t^{\dag}}^0 - \beta_{t,I_t^{\dag}}^0 \Vert x_t \Vert_{\left(V_{t,I_t^{\dag}}^0  \right)^{-1}} \le \min_i \langle x_t, \theta_i \rangle $ and Assumption~\ref{asp:nonworst}. We also have
\begin{equation}
\begin{aligned}
 (1-\alpha) x_t^T\theta_K + \alpha x_t^T \theta_{I_t^{\dag}}- (1-\alpha) x_t^T\theta_{K} = \alpha x_t^T \theta_{I_t^{\dag}} \ge 0. 
\end{aligned}
\end{equation}

\textbf{Case 2}:  when $\langle x_t,\hat{\theta}_{t,K}^0 \rangle \ge  \langle x_t,\hat{\theta}_{t,I_t^{\dag}}^0 \rangle > (1-2\alpha)\langle x_t,\hat{\theta}_{t,K}^0\rangle$ and $\epsilon_t = 1/2$, we have
\begin{equation}
    \mathbb{E}[r_{t,I_t^0}|F_{t-1}, I_t] = \frac{1}{2}\langle x_t, \theta_K \rangle +\frac{1}{2}\langle x_t,\theta_{I_t^{\dag}} \rangle.
\end{equation}
Then, by Lemma~\ref{lem:cboftar},
\begin{equation} \label{eq:case2}
\begin{aligned}
 &\frac{1}{2}( x_t^T\theta_K + x_t^T \theta_{I_t^{\dag}})- (1-\alpha) x_t^T\theta_{K} \\
= & \frac{1}{2}\left(  x_t^T \theta_{I_t^{\dag}} - (1-2\alpha) x_t^T\theta_{K}\right)\\
  \le & \frac{1}{2}\left(  x_t^T \hat{\theta}_{t,I_t^{\dag}}^0 +  \beta_{t,I_t^{\dag}}^0 \Vert x_t \Vert_{\left(V_{t,I_t^{\dag}}^0  \right)^{-1}} - (1-2\alpha) x_t^T\theta_{K}\right)\\
  \le  & \beta_{t,I_t^{\dag}}^0 \Vert x_t \Vert_{\left(V_{t,I_t^{\dag}}^0  \right)^{-1}}\nonumber
\end{aligned}
\end{equation}
where the last inequality is obtained by $ x_t^T \hat{\theta}_{t,I_t^{\dag}}^0 - \beta_{t,I_t^{\dag}}^0 \Vert x_t \Vert_{\left(V_{t,I_t^{\dag}}^0  \right)^{-1}} \le \min_i \langle x_t, \theta_i \rangle $ and  Assumption~\ref{asp:nonworst}.

In addition, by Lemma~\ref{lem:cboftar},
\begin{equation}
\begin{aligned}
 &\frac{1}{2}( x_t^T\theta_K + x_t^T \theta_{I_t^{\dag}})- (1-\alpha) x_t^T\theta_{K} \\
  \ge & \frac{1}{2}\left(  x_t^T \hat{\theta}_{t,I_t^{\dag}}^0 -  \beta_{t,I_t^{\dag}}^0 \Vert x_t \Vert_{\left(V_{t,I_t^{\dag}}^0  \right)^{-1}}  - (1-2\alpha)\left( x_t^T\hat{\theta}_{t,K}^0 + \beta_{t,K}^0 \Vert x_t \Vert_{\left(V_{t,K}^0  \right)^{-1}} \right)\right)\\
  \ge  & -\frac{1}{2} \beta_{t,I_t^{\dag}}^0 \Vert x_t \Vert_{\left(V_{t,I_t^{\dag}}^0  \right)^{-1}} - \frac{1}{2}(1 -2\alpha ) \beta_{t,K}^0 \Vert x_t \Vert_{\left(V_{t,K}^0  \right)^{-1}} .
\end{aligned}
\end{equation}

\textbf{Case 3}: when $0 \le  \langle x_t,\hat{\theta}_{t,I_t^{\dag}}^0 \rangle \le (1-2\alpha)\langle x_t,\hat{\theta}_{t,K}^0\rangle$ and $1/2 \le \epsilon_t \le 1 - \alpha $, we have
\begin{equation}
    \mathbb{E}[r_{t,I_t^0}|F_{t-1}, I_t] = \epsilon_t \langle x_t, \theta_K \rangle + (1-\epsilon_t)\langle x_t,\theta_{I_t^{\dag}} \rangle.
\end{equation}
We can find that
\begin{equation}
\begin{aligned}
& \epsilon_t \langle x_t, \theta_K \rangle + (1-\epsilon_t)\langle x_t,\theta_{I_t^{\dag}} \rangle - (1-\alpha) \langle  x_t,  \theta_K \rangle\\
= & \epsilon_t (\langle x_t, \theta_K \rangle -\langle x_t,\theta_{I_t^{\dag}} \rangle) + \langle x_t,\theta_{I_t^{\dag}} \rangle - (1-\alpha) \langle  x_t,  \theta_K \rangle\\
=&\epsilon_t (\langle  x_t, \hat{\theta}_{t,K}^0 \rangle - \langle x_t, \hat{\theta}_{t,I_t^{\dag}}^0  \rangle )+  \langle x_t,\theta_{I_t^{\dag}} \rangle - (1-\alpha) \langle  x_t,  \theta_K \rangle\\
&+ \epsilon_t (\langle x_t, \hat{\theta}_{t,I_t^{\dag}}^0  \rangle - \langle x_t,\theta_{I_t^{\dag}} \rangle) + \epsilon_t (\langle x_t, \theta_K \rangle - \langle  x_t, \hat{\theta}_{t,K}^0 \rangle)\\
=&(1-\alpha) \langle x_t, \hat{\theta}_{t,K}^0 \rangle - \langle x_t,\hat{\theta}_{t,I_t^{\dag}}^0 \rangle +  \langle x_t,\theta_{I_t^{\dag}} \rangle \\
&- (1-\alpha) \langle  x_t,  \theta_K \rangle + \epsilon_t (\langle x_t, \hat{\theta}_{t,I_t^{\dag}}^0  \rangle - \langle x_t,\theta_{I_t^{\dag}} \rangle)  + \epsilon_t (\langle x_t, \theta_K \rangle - \langle  x_t, \hat{\theta}_{t,K}^0 \rangle) \\
=&(1- \alpha - \epsilon_t) \left(\langle  x_t, \hat{\theta}_{t,K}^0 \rangle  -\langle x_t, \theta_K \rangle \right)  + (1- \epsilon_t)  \left(\langle  x_t, \hat{\theta}_{t,I_t^{\dag}}^0 \rangle  -\langle x_t, \theta_{I_t^{\dag}} \rangle \right),
\end{aligned}
\end{equation}
which is equivalent to 
\begin{equation}
\begin{aligned}
& \left|\mathbb{E}[r_{t,I_t^0}|F_{t-1}, I_t] - (1-\alpha) \langle  x_t,  \theta_K \rangle \right|\\
\le &(1- \alpha - \epsilon_t) \beta_{t,K}^0 \Vert x_t \Vert_{\left(V_{t,K}^0  \right)^{-1}}  \\
& + (1- \epsilon_t) \beta_{t,I_t^{\dag}}^0 \Vert x_t \Vert_{\left(V_{t,I_t^{\dag}}^0  \right)^{-1}} .
\end{aligned}
\end{equation}

\textbf{Case 4}: when $\langle x_t,\hat{\theta}_{t,I_t^{\dag}}^0 \rangle < 0$ and $\epsilon_t = 1 - \alpha $, we have
\begin{equation}
    \mathbb{E}[r_{t,I_t^0}|F_{t-1}, I_t] = (1 - \alpha) \langle x_t, \theta_K \rangle + \alpha \langle x_t,\theta_{I_t^{\dag}} \rangle.
\end{equation}
Then, by Lemma~\ref{lem:cboftar},
\begin{equation} 
\begin{aligned}
& (1-\alpha)  x_t^T\theta_K + \alpha x_t^T \theta_{I_t^{\dag}}- (1-\alpha) x_t^T\theta_{K} \\
=  &  \alpha x_t^T \theta_{I_t^{\dag}} \\
\le & \alpha x_t^T \hat{\theta}_{t,I_t^{\dag}}^0  +  \alpha  \beta_{t,I_t^{\dag}}^0 \Vert x_t \Vert_{\left(V_{t,I_t^{\dag}}^0  \right)^{-1}}   \\
  \le  & \alpha  \beta_{t,I_t^{\dag}}^0 \Vert x_t \Vert_{\left(V_{t,I_t^{\dag}}^0  \right)^{-1}} ,
\end{aligned}
\end{equation}
where the last inequality is obtained by $ x_t^T \hat{\theta}_{t,I_t^{\dag}}^0 - \beta_{t,I_t^{\dag}}^0 \Vert x_t \Vert_{\left(V_{t,I_t^{\dag}}^0  \right)^{-1}} \le \min_i \langle x_t, \theta_i \rangle $ and Assumption~\ref{asp:nonworst}. We also have
\begin{equation}
\begin{aligned}
 (1-\alpha) x_t^T\theta_K + \alpha x_t^T \theta_{I_t^{\dag}}- (1-\alpha) x_t^T\theta_{K} = \alpha x_t^T \theta_{I_t^{\dag}} \ge 0. 
\end{aligned}
\end{equation}

Combining these four cases, we have
\begin{equation}
\begin{aligned}
& \left|\mathbb{E}[r_{t,I_t^0}|F_{t-1}, I_t] - (1-\alpha) \langle  x_t,  \theta_K \rangle \right|\\
\le & (1-\alpha)\beta_{t,K}^0 \Vert x_t \Vert_{\left(V_{t,K}^0  \right)^{-1}}  + (1+\alpha) \beta_{t,I_t^{\dag}}^0 \Vert x_t \Vert_{\left(V_{t,I_t^{\dag}}^0  \right)^{-1}} .
\end{aligned}
\end{equation}

\section{Proof of Lemma~\ref{lem:bba}} \label{prfbba}
From Section~\ref{proofl1}, we have, for any arm  $i \ne K$,
\begin{equation}  \label{eq:badivide}
\begin{aligned}
    &|x_t^T\hat{\theta}_{t,i}-x_t^T (1-\alpha) \theta_K|\\
    \le&\left|\sum_{k  \in \tau_i(t-1)} x_t^T V_{t,i}^{-1} x_{k}\left( x_{k}^T\theta_{I_k^0}-(1-\alpha) x_{k}^T\theta_K\right)\right| \\
    & + \left|\sum_{k  \in \tau_i(t-1)} x_t^T V_{t,i}^{-1} x_{k} \eta_k \right| + \left|\lambda x_t^T V_{t,i}^{-1} (1-\alpha) \theta_K \right| \\
    \le&\left|\sum_{k  \in \tau_i(t-1)} x_t^T V_{t,i}^{-1} x_{k}\left( x_{k}^T\theta_{I_k^0}- \epsilon_k\langle x_k,\theta_{K} \rangle  - (1-\epsilon_k)\langle x_k,\theta_{I_k^{\dag}} \rangle\right)\right| \\
    & + \left|\sum_{k  \in \tau_i(t-1)} x_t^T V_{t,i}^{-1} x_{k}\left(  \epsilon_k\langle x_k,\theta_{K} \rangle + (1-\epsilon_k)\langle x_k,\theta_{I_k^{\dag}} \rangle-(1-\alpha) x_{k}^T\theta_K\right)\right| \\
    &+ \left|\sum_{k  \in \tau_i(t-1)} x_t^T V_{t,i}^{-1} x_{k} \eta_k \right| + \left|\lambda x_t^T V_{t,i}^{-1} (1-\alpha) \theta_K \right|.
\end{aligned}
\end{equation}

Since the mean rewards are bounded and the rewards are generated independently, we have $0 \le \left| x_{k}^T\theta_{I_k^0} - \epsilon_k\langle x_k,\theta_{K} \rangle- (1-\epsilon_k)\langle x_k,\theta_{I_k^{\dag}} \rangle \right| \le LS$ and 
$\mathbb{E}[ x_{k}^T\theta_{I_k^0} | F_{k-1} ]= \epsilon_k\langle x_k,\theta_{K} \rangle + (1-\epsilon_k)\langle x_k,\theta_{I_k^{\dag}} \rangle $. 

Then $\left\{ x_t^T V_{t,i}^{-1} x_{k}\left( x_{k}^T\theta_{I_k^0} - \mathbb{E}[ x_{k}^T\theta_{I_k^0} | F_{k-1} ] \right)\right\}_{k \in \tau_i(t-1)}$ is also a bounded martingale difference sequence w.r.t the filtration $\{F_k\}_{k \in \tau_i(t-1)}$. By following the steps in Section~\ref{proofl1}, we have, with probability $1-\delta$, for any arm $i$ and any round $t$, 
\begin{equation}
\begin{aligned}
     &\left|\sum_{k  \in \tau_i(t-1)} x_t^T V_{t,i}^{-1} x_{k}\left( x_{k}^T\theta_{I_k^0}- \mathbb{E}[ x_{k}^T\theta_{I_k^0} | F_{k-1} ] \right)\right| \\
     \le & LS\sqrt{\frac{1}{2}\log\left(\frac{2KT}{\delta}\right)  } \Vert x_t \Vert_{V_{t,i}^{-1}}.
\end{aligned}
\end{equation}

From~\eqref{fact1} in Section \ref{proofl1}, we have 
\begin{equation} 
\begin{aligned}
\Vert x_t \Vert_{V_{t,i}^{-1}}^2  \ge \sum_{k  \in \tau_i(t-1)}(x_t^T V_{t,i}^{-1} x_{k})^2.
\end{aligned}
\end{equation}
Then, the second item of the right hand side of~\eqref{eq:badivide} can be upper bounded by
\begin{equation}
\begin{aligned}
     &\left|\sum_{k  \in \tau_i(t-1)} x_t^T V_{t,i}^{-1} x_{k}\left(  \epsilon_k \langle x_k,\theta_{K} \rangle + (1-\epsilon_k)\langle x_t,\theta_{I_k^{\dag}} \rangle \right.\right.\\
     & \left.\left.-(1-\alpha) x_{k}^T\theta_K\right)\right|\\
     \le &  \sqrt{\sum_{k  \in \tau_i(t-1)}\left( \mathbb{E}[r_{k,I_k^0}|F_{k-1}, I_k]-(1-\alpha) x_{k}^T\theta_K\right)^2}  \sqrt{\sum_{k  \in \tau_i(t-1)}(x_t^T V_{t,i}^{-1} x_{k})^2}\\
     \le & \left(\sum_{k  \in \tau_i(t-1)}\left( (1-\alpha)\beta_{k,K}^0 \Vert x_k \Vert_{\left(V_{k,K}^0  \right)^{-1}}  + (1+\alpha) \beta_{k,I_k^{\dag}}^0 \Vert x_k \Vert_{\left(V_{k,I_k^{\dag}}^0  \right)^{-1}}\right)^2\right)^{\frac{1}{2}} \Vert x_t \Vert_{V_{t,i}^{-1}},
\end{aligned}
\end{equation}
where the first inequality is obtained from Cauchy-Schwarz inequality, the second inequality is obtained from Lemma~\ref{lem:difbtw} and~\eqref{fact1}. 

In addition, by the fact that $(a+b)^2 \le 2a^2+2b^2$ for any real number, we have 
\begin{equation} \label{eq:factsqrt}
\begin{aligned}
& \sum_{k  \in \tau_i(t-1)}\left( (1-\alpha)\beta_{k,K}^0 \Vert x_k \Vert_{\left(V_{k,K}^0  \right)^{-1}}  + (1+\alpha) \beta_{k,I_k^{\dag}}^0 \Vert x_k \Vert_{\left(V_{k,I_k^{\dag}}^0  \right)^{-1}}\right)^2 \\
\le & \sum_{k  \in \tau_i(t-1)} 2\left( (1-\alpha)\beta_{k,K}^0 \Vert x_k \Vert_{\left(V_{k,K}^0  \right)^{-1}} \right)^2 + \sum_{k  \in \tau_i(t-1)} 2\left( (1+\alpha) \beta_{k,I_k^{\dag}}^0 \Vert x_k \Vert_{\left(V_{k,I_k^{\dag}}^0  \right)^{-1}}\right)^2.
\end{aligned}
\end{equation}

Here, we use Lemma 11 from \cite{abbasi2011improved} and get, for any arm $i$,
\begin{equation}
\begin{aligned}
\sum_{k \in \tau_i^{\dag}(t-1) } \Vert x_k \Vert_{(V_{k,i}^0)^{-1}}^2 & \le 2d\log\left (1+\frac{N_i(t) L^2}{d \lambda}\right)  \\
& \le 2d\log \left(1+\frac{t L^2}{d \lambda}\right).
\end{aligned}
\end{equation}

By the fact that $\sum_{i}\tau_i(t-1) = \tau_K^{\dag}(t-1)$, and $\sum_{i \ne K} \tau_i(t-1) = \sum_{i \ne K} \tau_i^{\dag}(t-1)$, we have, for any arm $i$, $\tau_i(t-1) \subseteq \tau_K^{\dag}(t-1)$, and $ \tau_i(t-1) \subseteq \sum_{j \ne K} \tau_j^{\dag}(t-1)$. Thus,
\begin{equation} \label{eq:sumdagK}
\begin{aligned}
\sum_{k  \in \tau_i(t-1)} \Vert x_k \Vert_{\left(V_{k,K}^0  \right)^{-1}}^2 & \le \sum_{k \in \tau_K^{\dag}(t-1) } \Vert x_k \Vert_{(V_{k,K}^0)^{-1}}^2 \\
& \le  2d\log \left(1+\frac{t L^2}{d \lambda}\right),
\end{aligned}
\end{equation}
and
\begin{equation} \label{eq:sumdagother}
\begin{aligned}
\sum_{k  \in \tau_i(t-1)} \Vert x_k \Vert_{\left(V_{k,I_k^{\dag}}^0 \right)^{-1}}^2 &  \le \sum_{i \ne K} \sum_{k \in \tau_i^{\dag}(t-1) } \Vert x_k \Vert_{(V_{k,i}^0)^{-1}}^2\\
& \le  2(K-1)d\log \left(1+\frac{t L^2}{d \lambda}\right).
\end{aligned}
\end{equation}

By combining~\eqref{eq:factsqrt},~\eqref{eq:sumdagK} and~\eqref{eq:sumdagother}, we have
\begin{equation}
\begin{aligned}
& \sum_{k  \in \tau_i(t-1)}\left( (1-\alpha)\beta_{k,K}^0 \Vert x_k \Vert_{\left(V_{k,K}^0  \right)^{-1}}  + (1+\alpha) \beta_{k,I_k^{\dag}}^0 \Vert x_k \Vert_{\left(V_{k,I_k^{\dag}}^0  \right)^{-1}}\right)^2 \\
\le & \sum_{k  \in \tau_i(t-1)} 2\left( \beta_{k,K}^0 \Vert x_k \Vert_{\left(V_{k,K}^0  \right)^{-1}} \right)^2  + \sum_{k  \in \tau_i(t-1)} 2\left( 2 \beta_{k,I_k^{\dag}}^0 \Vert x_k \Vert_{\left(V_{k,I_k^{\dag}}^0  \right)^{-1}}\right)^2\\
\le &  16d^2\left(1+\frac{(K-1) }{\alpha^2}\right) \left(\omega(t)+LS\sqrt{\frac{1}{2}\log\left(\frac{2KT}{\delta}\right)} \right)^2 \log \left(1+\frac{t L^2}{d \lambda}\right) \\
\le &  \frac{16d^2K }{\alpha^2} \left(\omega(t)+LS\sqrt{\frac{1}{2}\log\left(\frac{2KT}{\delta}\right)} \right)^2 \log \left(1+\frac{t L^2}{d \lambda}\right).
\end{aligned}
\end{equation}

In summary, we have
\begin{equation}
\begin{aligned}
   & |x_t^T\hat{\theta}_{t,i}-x_t^T (1 - \alpha) \theta_K| \\
 \le    & \left( 1+ \frac{4d }{\alpha}\sqrt{K\log \left(1+\frac{t L^2}{d \lambda}\right)} \right)  \left(  \omega(N_i(t))+LS\sqrt{\frac{1}{2}\log\left(\frac{2KT}{\delta}\right)}\right) \Vert x_t \Vert_{V_{t,i}^{-1}} .
\end{aligned}
\end{equation}

\section{Proof of Theorem~\ref{thm:bbacost}} \label{prooft2}
For round $t$ and context $x_t$, if LinUCB pulls arm $i \ne K$, we have
\begin{equation}
\begin{aligned}
x_t^T \hat{\theta}_{t,K} + \beta_{t,K} \sqrt{x_t^T V_{t,K}^{-1} x_t} \le x_t^T \hat{\theta}_{t,i} + \beta_{t,i} \sqrt{x_t^T V_{t,i}^{-1} x_t}.\nonumber
\end{aligned}
\end{equation}
In this case, $\beta_{t,i} = \omega(N_i(t)) = \sqrt{\lambda}S + R\sqrt{2\log\frac{K}{\delta}+d\log\left( 1+ \frac{L^2 N_i(t)}{\lambda d} \right)}$.

Since the attacker does not attack the target arm, the confidence bound of arm $K$ does not change and $x_t^T \theta_K \le x_t^T \hat{\theta}_{t,K} + \beta_{t,K} \sqrt{x_t^T V_{t,K}^{-1} x_t} $ holds with probability $1-\frac{\delta}{K}$.

Thus, by Lemma 2,
\begin{equation}
\begin{aligned}
x_t^T \theta_K \le & x_t^T \hat{\theta}_{t,i} + \beta_{t,i} \sqrt{x_t^T V_{t,i}^{-1} x_t} \\
\le&
x_t^T (1-\alpha) \theta_K + \omega(N_i(t))  \Vert x_t \Vert_{V_{t,i}^{-1}}   \\
&+ \left( 1+ \frac{4d }{\alpha}\sqrt{K\log \left(1+\frac{t L^2}{d \lambda}\right)} \right) \left(  \omega(N_i(t))+LS\sqrt{\frac{1}{2}\log\left(\frac{2KT}{\delta}\right)}\right) \Vert x_t \Vert_{V_{t,i}^{-1}} .
\end{aligned}
\end{equation}

By multiplying both sides by $\mathbbm{1}_{\{ I_t = i \}}$ and summing over rounds, we have
\begin{equation}
\begin{aligned}
 & \sum_{k=1}^t \mathbbm{1}_{\{ I_k = i \}} \alpha x_k^T \theta_K \\
\le  & \sum_{k=1}^t \mathbbm{1}_{\{ I_k = i \}} \left( 2+ \frac{4d }{\alpha}\sqrt{K\log \left(1+\frac{k L^2}{d \lambda}\right)} \right) \\
& \left(  \omega(N_k(t))+LS\sqrt{\frac{1}{2}\log\left(\frac{2KT}{\delta}\right)}\right) \Vert x_k \Vert_{V_{k,i}^{-1}} .
\end{aligned}
\end{equation}

Here, we use Lemma 11 from \cite{abbasi2011improved} and get 
\begin{equation}
\begin{aligned}
\sum_{k=1}^t \mathbbm{1}_{\{ I_k = i \}}  \Vert x_k \Vert_{V_{k,i}^{-1}}^2 & \le 2d\log \left(1+\frac{N_i(t) L^2}{d \lambda}\right) \\
& \le 2d\log \left(1+\frac{t L^2}{d \lambda}\right).
\end{aligned}
\end{equation}

According to $ \sum_{k=1}^t \mathbbm{1}_{\{ I_k = i \}} \Vert x_k \Vert_{V_{k,i}^{-1}} \le \sqrt{N_i(t) \sum_{k=1}^t \mathbbm{1}_{\{ I_k = i \}}  \Vert x_k \Vert_{V_{k,i}^{-1}}^2}$, we have
\begin{equation}
\begin{aligned}
\sum_{k=1}^t \mathbbm{1}_{\{ I_k = i \}}  \Vert x_k \Vert_{V_{k,i}^{-1}}^2 \le \sqrt{N_i(t)2d\log \left(1+\frac{t L^2}{d \lambda}\right)}.
\end{aligned}
\end{equation}

Thus, we have
\begin{equation}
\begin{aligned}
 & \sum_{k=1}^t \mathbbm{1}_{\{ I_k = i \}} \alpha x_k^T \theta_K \\
\le  & \sqrt{N_i(t)2d\log \left(1+\frac{t L^2}{d \lambda}\right)} \left( 2+ \frac{4d }{\alpha}\sqrt{K\log \left(1+\frac{t L^2}{d \lambda}\right)} \right)\\
& \left(  \omega(t)+LS\sqrt{\frac{1}{2}\log\left(\frac{2KT}{\delta}\right)}\right),
\end{aligned}
\end{equation}
and
\begin{equation}
\begin{aligned}
 & N_i(t) = \sum_{k=1}^t \mathbbm{1}_{\{ I_k = i \}}  \\
\le  & \frac{2d}{(\alpha \gamma)^2} \log \left(1+\frac{t L^2}{d \lambda}\right) \left( 2+ \frac{4d }{\alpha}\sqrt{K\log \left(1+\frac{t L^2}{d \lambda}\right)} \right)^2\\
& \left(  \omega(t)+LS\sqrt{\frac{1}{2}\log\left(\frac{2KT}{\delta}\right)}\right)^2,
\end{aligned}
\end{equation}
where $\gamma = \min_{x \in \mathcal{D}} \langle x,\theta_K \rangle$.

\end{document}